\documentclass[10pt]{article}

\usepackage{arxiv}

\usepackage[utf8]{inputenc} 
\usepackage[T1]{fontenc}    
\usepackage{hyperref}       
\usepackage{url}            
\usepackage{booktabs}       
\usepackage{amsfonts}       
\usepackage{nicefrac}       
\usepackage{microtype}      
\usepackage{lipsum}
\usepackage{blindtext, graphicx}

\title{Feature Selection Using Reinforcement Learning}

\author{
  Sali Rasoul \\ 
  Systems and Information Engineering\\
  University of Virginia\\
  Charlottesville, VA - 22904\\
  \texttt{rs8wa@virginia.edu} \\
   \And
 Sodiq Adewole \\
  Systems and Information Engineering\\
  University of Virginia\\
  Charlottesville, VA - 22904\\
  \texttt{soa2wg@virginia.edu} \\
  
  \And
   Alphonse Akakpo \\
  Systems and Information Engineering\\
  University of Virginia\\
  Charlottesville, VA - 22904\\
  \texttt{anacy2@virginia.edu} \\
  
}

\begin{document}
\maketitle

\begin{abstract}
With the decreasing cost of data collection, the space of variables or features that can be used to characterize a particular predictor of interest continues to grow exponentially. Therefore, identifying the most characterizing features that minimizes the variance without jeopardizing the bias of our models is critical to successfully training a machine learning model. In addition, identifying such features is critical for interpretability, prediction accuracy and optimal computation cost. While statistical methods such as subset selection, shrinkage, dimensionality reduction have been applied in selecting the best set of features, some other approaches in literature have approached feature selection task as a search problem where each state in the search space is a possible feature subset. In this paper, we solved the feature selection problem using Reinforcement Learning. Formulating the state space as a Markov Decision Process (MDP), we used Temporal Difference (TD) algorithm to select the best subset of features. Each state was evaluated using a robust and low cost classifier algorithm which could handle any non-linearities in the dataset.
\end{abstract}


\section{Introduction}
In statistical machine learning, least squares is used for model fitting and prediction error minimization, alternative fitting procedure such as subset selection, shrinkage and dimension reduction are also used to improve prediction accuracy as well as improve model interpretability. As we are able collect more and more data \cite{8735621}, the number of features for a predictor of interest has continued to grow exponentially and minimizing bias and effect of noise in our models is particularly critical if $n$, the number of observations is not much larger than $p$, the number of features. Low bias, which is also achieved with increased complexity of the model and low variance which is achievable with increased data size poses the double sided-problem of bias-variance trade-off. Approaches such as subset selection and dimension reduction are used to solve the problem of model complexity and generalized prediction accuracy which are both aimed at minimizing the bias and variance of the learned hypothesis. Three main approaches of Features Selection, which is a combinatorial optimization problem are filtering (scoring), wrapper and embedded approaches.

\subsection{Feature Selection Approaches}
Feature selection aims to reduce generalization error on learned hypothesis by removing noise (filtering out irrelevant features) or considering simpler hypothesis spaces (filtering out redundant features). Feature Selection reduces dimension of data in contrast to feature extraction methods \cite{adewole2020deep}. In feature selection approaches, the amount of information in every feature remains intact while the feature space is optimally reduced according to a certain criterion. Categories of Features selection algorithms are;\\

\textbf{Filtering approaches} this algorithm proceed by independently ranking features using some scoring function and selecting top-ranked ones without reference to the chosen predictor. Typical scoring functions include measuring correlation coefficient between the features and the class values using student's t-test or Analysis of Variance or Fisher test and then selecting the best set of features based on the score. This score function discards feature inter-dependencies which constitute the main limitation of the approach. Other information criteria could also be used alongside the correlation  coefficient as a scoring function between the features.\\

\textbf{Wrapper approaches} tackle the combinatorial optimization problem by exploring the power set of the feature set and measuring the generalization error of all considered subsets. This approach utilizes the machine learning algorithm of interest to score the subsets of features according to their predictive power. However, navigating this power set space with large number of features while dealing with exploration and exploitation dilemma could be very costly. On the one had, exploitation (myopic search) is bound to end up in a local optimum. Hill-climbing approaches lead to useful methods named correlation-based feature selection (CFS) \cite{hall2000correlation}. Combining global and local search operators \cite{boulle2007compression}  mixing forward and backward selection \cite{zhang2008adaptive}, using look-ahead or making large steps in the lattice \cite{margaritis2009toward}, among many other heuristics have been proposed to address the exploration vs exploitation dilemma. Wrappers are considered as the best, among the other methods, however, the exhaustive search of the power set of the feature set makes it highly complex and the high computational cost could potentially impact the benefits.\\

\textbf{Embedded Approaches} combine learning and feature selection through prior or posterior regularization \cite{gaudel2010feature}. \cite{james2013introduction} pioneered the use of $L_{1}$ regularization called Least absolute shrinkage and selection operator (Lasso) approach. They exploit the learned hypothesis and incorporate feature selection as part of the training process. They converge faster to a solution by avoiding retraining a predictor from scratch for every feature subset.\\Decision trees such as CART have a built-in feature selection mechanism. A famous approach in this field uses random forests to compute feature scores. The main weakness of these Feature Selection approach is in feature redundancy. Recent deep learning methods \cite{adewole2020deep, adewole2021lesion2vec, sali2020hierarchical, adewole2020dialogue} have demonstrated more advanced capability on automatic feature selection, however, they require large dataset and are sample inefficient. 

\section{Feature Selection as Reinforcement Learning}
\ In this paper, we try to solve feature selection problem using Reinforcement Learning approach where the state space comprises of all possible subsets of the features and an action is any feature that is included in the model. The action space for every state is determined by the number of features not already included in the model. This way, we are able to reduce the search space for the next best action, avoid overly sparse state space and increase the speed of computation. The reward function is determined by the scored accuracy of the machine learning algorithm used in evaluating the predictive power of the current state (current set of features). The reward is the difference in accuracy value for the current state and the next state after taking an action of including an additional feature in the model. The average of collected rewards for that feature in several iterations is considered its final score and this is used in evaluate the next best action in every state. This difference between the values of two consecutive states indicates the effectiveness of the corresponding feature that causes the transition. The average of these values in several iterations is defined as the Average of Reward of the feature. We define the reward of action $f$ as follows

\begin{equation}
    Reward_{f} = [Accuracy]_{t+1} - [Accuracy]_{t} \hspace{1.5cm}
\end{equation}

We used Temporal Difference (TD) algorithm in evaluating state value based on the reward obtained and the value of the state when it was previously visited.

This problem is formulated to handle high dimensional features space and robust enough for any non-linear relationship between the predictors and the response variable. We  used Support Vector Machines (SVM) classifier for evaluating the state value for each selected subset. SVM behave robustly over a variety of different learning tasks. They are also fully automatic, eliminating the need for manual parameter tuning classifier \cite{joachims1998text}. SVM classifiers also performs well in large dimensional space and very robust for non-linear classification problems.

\section{Problem Formulation}

We first formulate the problem of feature selection as a Markov Decision Process (MDP) within the Feature space. Let $\psi$ stand for the set of all features and $F$ be a subset of $\psi$. The state space is the power set of $\psi$. An action could be the selection of an unseen feature ($f$ from $\psi$) and is regarded as transition from one state to another. Each iteration starts with a random set of features and proceeds by adding one selected feature to the previous set until all features are added.\\
Selecting a feature $f$ from a set of all features in every state to reach the optimal value of the state after several iterations determines the optimal policy $\pi^*$. This leads to selecting the best reward based on the following:

\begin{equation}
    \pi^*(F) = argmax_{f\in \psi /F}(rewards) \hspace{2cm}
\end{equation}

Given that our goal is to maximize the generalization accuracy and by this minimize the error of the learned hypothesis from the training set, the best policy can be shown to be

\begin{equation}
    \pi* = argmin_{\pi} Error(A(F_{\pi})) \hspace{2cm}
\end{equation}

where $A$ is the learned hypothesis which in our case is the model output of the SVM classifier.

\section{Proposed Reinforcement Learning Framework}
We define the state space as the power set of all the features $\psi$ and a reward for an action of selecting any feature in the current state is given by (1) above.\\ 
With all states initialized to zero, in the TD method, to update the value of the current state, it is necessary to select the best action of this state and move to the next state instead of waiting until the final stage is reached as it is the case with monte carlo method. The target for the TD(0) method is to update the value of the current state as follows:

\begin{equation}
    V(S_t) \leftarrow V(S_t) + \alpha [r_{t+1} + \gamma V(S_{t+1}) - V(S_{t})] \hspace{0.5cm}
\end{equation}

where $V(S_{t})$ and $V(S_{t+1})$ are the values of the current and next state respectively, $r_t+1$ is the reward that is gained for an action of selecting feature $f$ in state $S_{t}$ to transition to a new state $S_{t+1}$. Other hyper-parameters $\alpha$ and $\gamma$ are constants between 0 and 1, where $\alpha$ is the learning rate used to control the rate of update and $\gamma$ is a discount factor to moderate the effect of observing the next state. When $\gamma$ is close to 0, this means the estimated value of the next state is worth as little as nothing and this is a shortsighted condition; when it tends to 1, next state estimated value is worth the full amount and this translates into a farsighted behaviour.\\

\subsection{Feature scoring}
To determine the next best action in a state, a measure is necessary to compare features. A criterion known as average of reward is introduced which is the average of rewards of each feature whenever it is selected in a state:

\begin{equation}
    AOR_f = Average\{V(F_t) - V(F_t+1)\} \hspace{1cm}
\end{equation}

where $V(F_t)$ and $V(F_t+1)$ are the assessment values of the current successor states and $f$ is the feature that causes a transfer from state $F_t$ to the next one $F_t+1$. In this criterion, the difference between the values of the current and the next state reveals the influence of the selected feature and the reward function is a Gaussian SVM in accordance with the features of the states, as mentioned already \cite{fard2013using}.

In calculating this criterion, there is no need to maintain all rewards of a feature, whenever it is selected, since it can be computed in an incremental manner [3]. To do so, a 2 by n array can be constructed at the beginning of the algorithm; where n is the number of features. Then each column belongs to a feature; the first row indicates the number of times that this feature is selected, and the second row shows the AOR value of the relevant feature. All of the array cells are initially set to zero, and each time a feature is selected the visiting number is incremented by 1. The AOR value can be updated incrementally as:

\begin{equation}
    (AOR_f)_{New} = [(k - 1)(AOR_f)_{Old} + V(F)]/k
\end{equation}

where $(AOR_f)_{New}$ and $(AOR_f)_{Old}$ are the current and previous values of $AOR$ for feature $f$ and $k$ is the number of times this feature has been selected.\\

\subsection{Feature selection}
We split the feature selection approach into two phases. At the initialization of the state evaluation, the agent has no knowledge of the environment and so has no experience of the measure of reward that can be received for any selected feature. This constrains the agent to first explore the environment by taking random actions and experiencing the reward. This reward for the selected action can then be saved into a look-up table where on subsequent visit, the agent can act to maximize the reward received in any given state. This allows the agent to trade-off exploration and exploitation. We also initialize every episode with a random selection of state, this will allow the agent to explore and visit state that has not been seen before. Two phases of feature selection are random based and bandit based phases which we further explain below; 

-\textit{the random based phase}: Whenever an unseen state is met, it is added to the tree as a new node and because of lack of experience about the state, random selection of features is again invoked as well as free exploration of state space.\\

-\textit{the epsilon-greedy phase}:This phase is used when the algorithm reaches a node that has been observed before, so the node preserves a set of performed actions(called repository) in transmitting to another state. In this stage, one of these actions that maximizes the prediction accuracy of the learned hypothesis could be selected with a high probability. 

\begin{equation}
    f^* = max_{f\in \psi^{'}} [AOR_{f}]
\end{equation}

where $\psi^{'}$ is the set of available features \textit{f} that is not already in the model for the current state and $AOR_{F}$ as described above is the average of the rewards that has been received in the past for selected feature \textit{f}.

This equation balances the exploration vs exploitation of the algorithm. While the first part of equation (5) allows action from the repository that has collected more average of rewards when transferring from this node to another to have more chance to be selected, actions that have received less selection will have low $t_{F,a}$ value and the second part of the equation will be larger. Thereby increasing exploration of other actions.\\
Features selection continues this way until all features have been selected, a stopping condition is reached  or a new node is reached at which point it must switch to the random-based phase.

\subsection{Trajectory of an Episode}
After we initialize every episode with random state selection, the agent proceeds through the episode by either random action selection or epsilon-greedy action selection. This dependent on whether this is the first visit to the state or subsequent visit. At the first visit, the agent randomly selects an action, while in subsequent visit, it acts greedily based on the $AOR$ value of the available actions in that state. The agent then sequentially add features and keeps evaluating the reward of the selected feature while also updating the $AOR$ table to keep track of the value of selected features over the course of the episode. The terminal state is reached when all features in $\psi$ have been included in the model.

\section{Experimentation and Results}
We ran this algorithm on three (3) datasets tuning different parameters; we selected datasets based on the number of features so as to apply our algorithm on large, medium and small feature space. We also considered scalability of this algorithm to large datasets in terms of number of observations and we chose a range of observations that spans approximately between 200 and 700. The following table shows properties of our selected datasets;\\

\begin{center}
\begin{tabular}{|p{0.7cm}|p{6.5cm}|p{1.5cm}|p{1.5cm}|}
 \hline
 \multicolumn{4}{|c|}{Experimental Datasets} \\
 \hline
 \textbf{S.No}. & \textbf{Dataset Name} & \textbf{Features} & \textbf{Samples}\\
 \hline
 1 & Australian & 14 & 690\\
 \hline
 2 & Breast Cancer Wisconsin (Prognostic) - WPBC  & 34 & 198\\
 \hline
 3 & Connectionist Bench & 60 &  208\\
 \hline
\end{tabular}
\end{center}

We see from the plot below that the state value increases to a maximum of around 3.5 around the $50^{th}$ episode.

\begin{figure}[ht]
    \centering
    \includegraphics[width=0.7\linewidth]{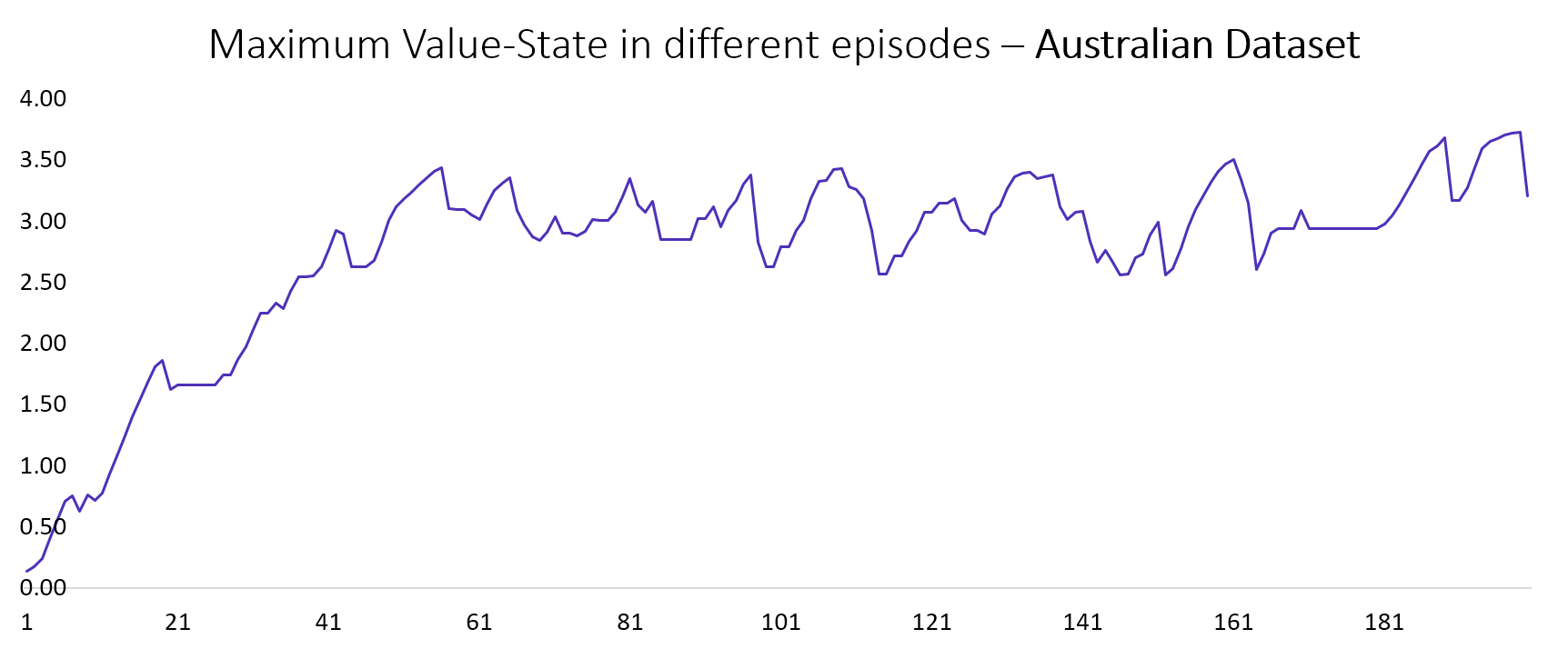}
    \caption{Maximum Value-State in Different Episodes - Australian Dataset}
    \label{fig:value_max}
\end{figure}

We also tuned different values of epsilon to investigate the effect on the number of states visited by the agent which informs us of the rate of exploration and hence update of state value function of the selected features. As see below, the number of states visited increases with increasing values of epsilon.\\

\begin{figure}[ht]
    \centering
    \includegraphics[width=0.7\linewidth]{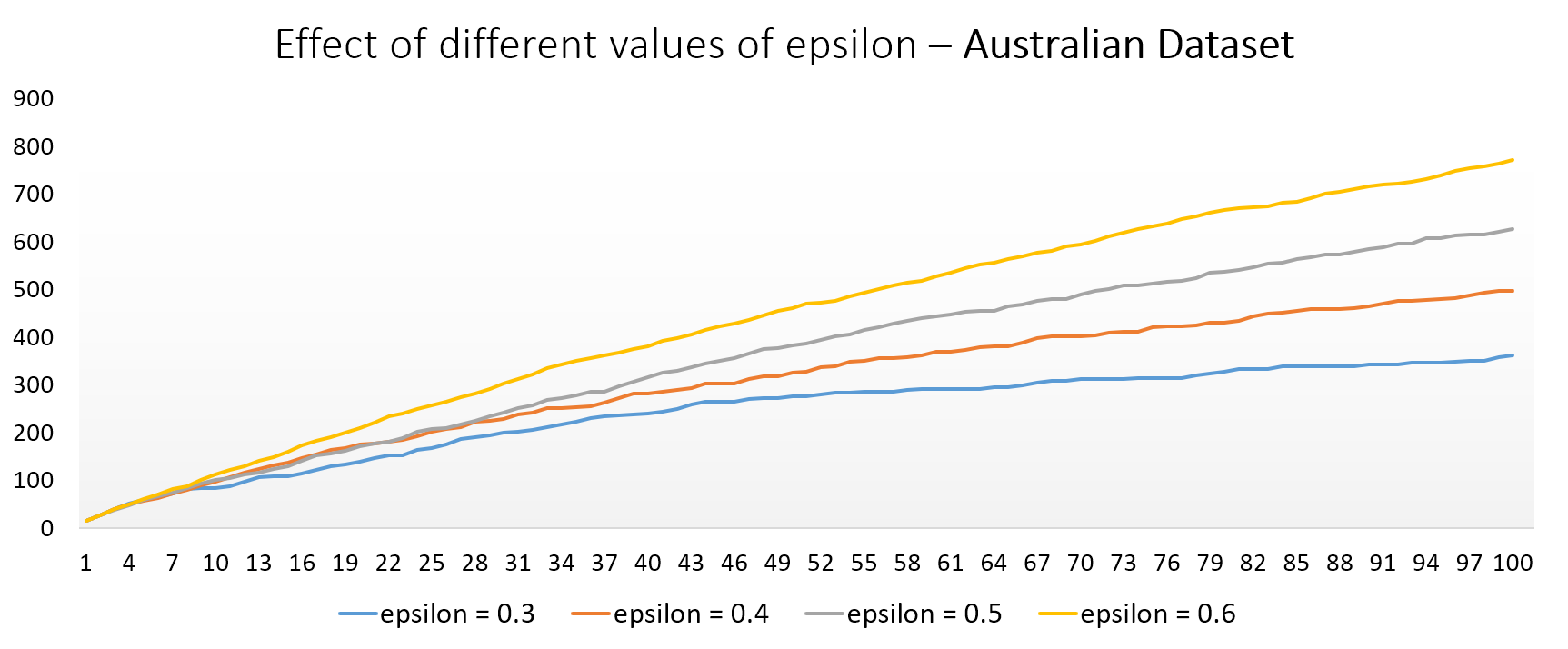}
    \caption{Effect of Different Values of Epsilon - Australian Dataset}
    \label{fig:epsilon}
\end{figure}

\vspace{0.5cm}
Further, the result of our experiment based on the wrapper based method of evaluating the accuracy of the selected subset shows;\\

\begin{center}
\begin{tabular}{|p{0.5cm}|p{6.5cm}|p{2.5cm}|}
 \hline
 \multicolumn{3}{|c|}{Results of Classification Accuracy} \\
 \hline
 No & Name & Accuracy (\%) \\
 \hline
 1 & Australian & $85.55 \pm 0.039$\\
 2 & Breast Cancer Wisconsin (Prognostic) - WPBC  & $76.29 \pm 0.007$\\
 3 & Connectionist Bench & $73.69 \pm 0.108$ \\
 \hline
\end{tabular}\\
\end{center}

For filter based evaluation of the result of experimentation on the dataset where we show ranking of the features using different methods on each of our experimental dataset.\\

\begin{figure}[ht]
    \begin{center}
        \centering
        \includegraphics[width=0.7\linewidth]{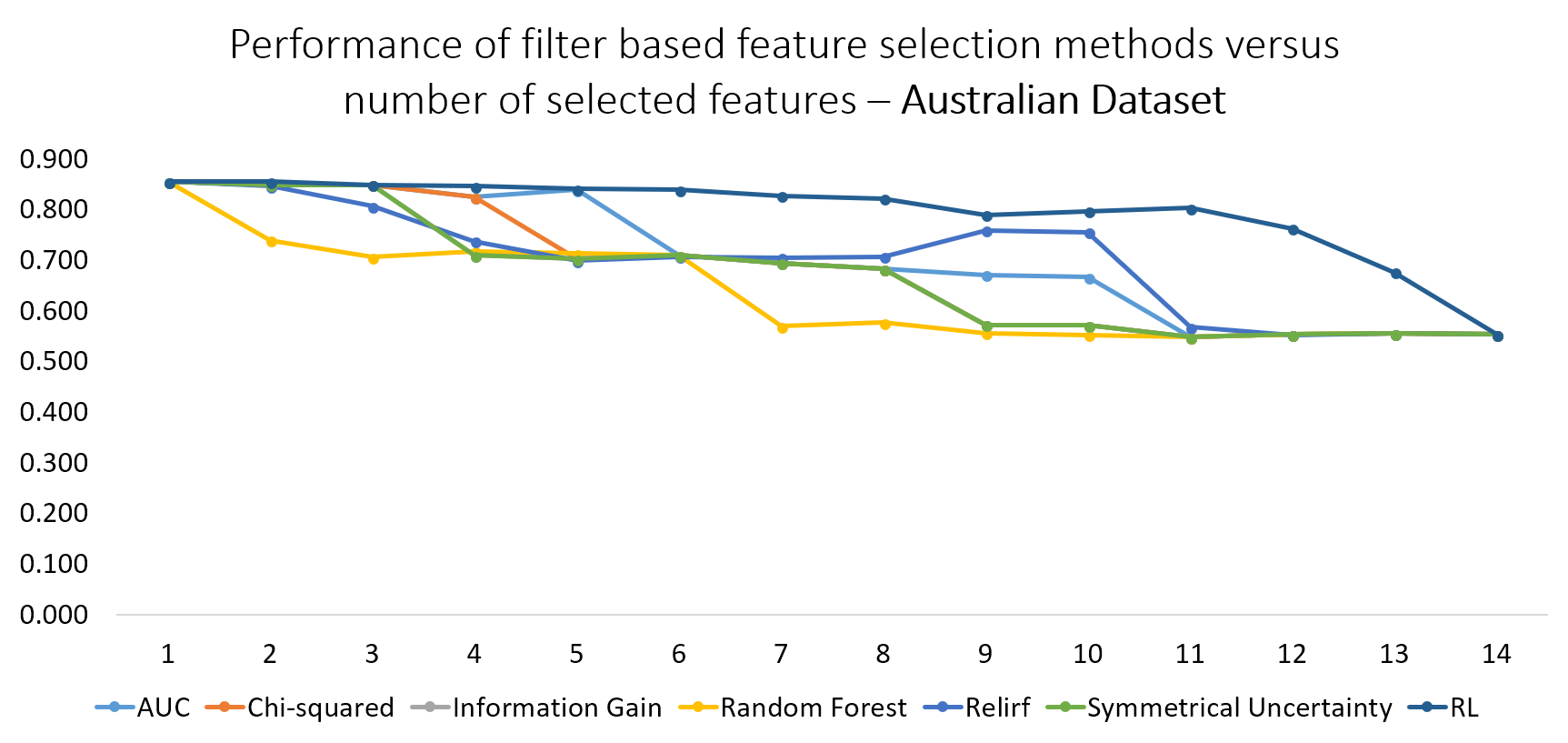}
        \caption{Performance of filter-based feature selection methods versus number of selected features - Australian Dataset}
        \label{fig:my_label}
    \end{center}
\end{figure}

\begin{figure}[ht]
    \centering
    \includegraphics[width=0.7\linewidth]{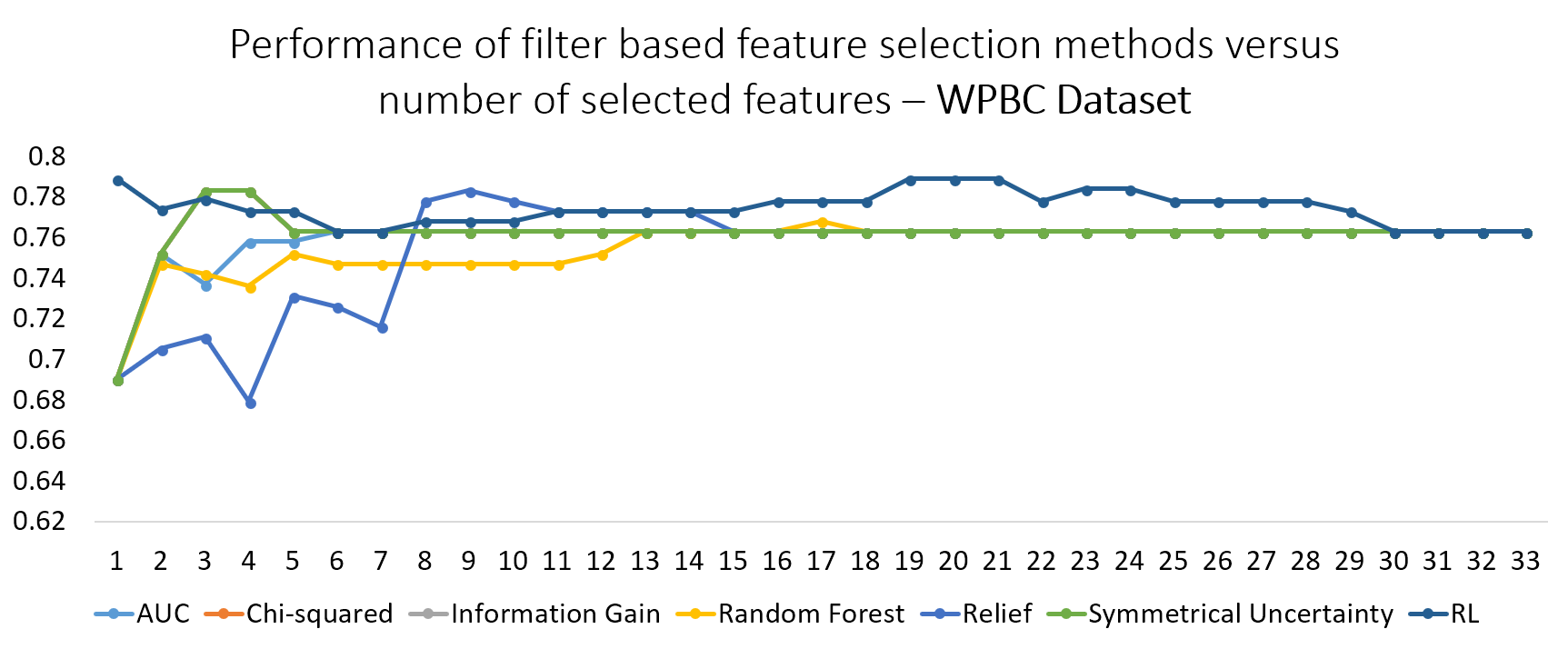}
    \caption{Performance of filter-based feature selection methods versus number of selected features - WPBC Dataset}
    \label{fig:wpbc}
\end{figure}

\begin{figure}[ht]
    \centering
    \includegraphics[width=0.7\linewidth]{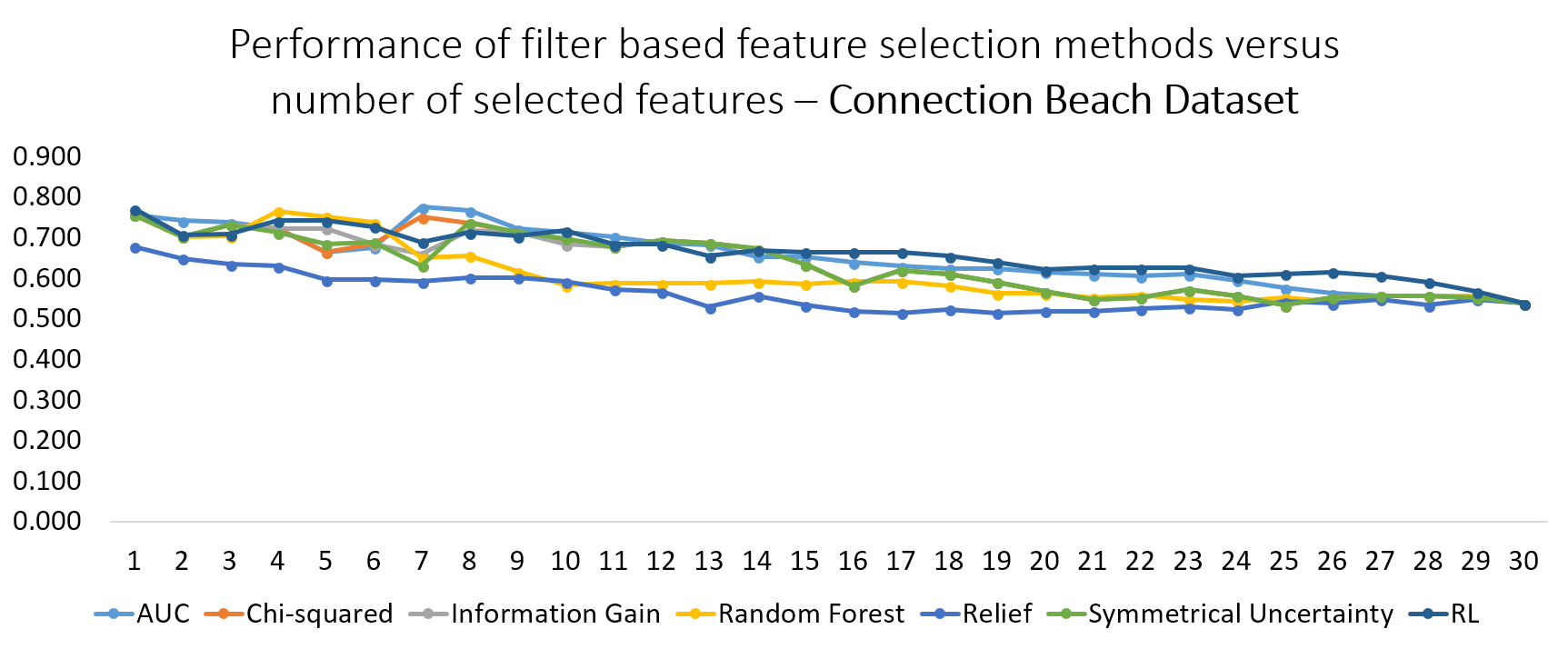}
    \caption{Performance of filter-based feature selection methods versus number of selected features - Connection Beach Dataset}
    \label{fig:}
\end{figure}

This results show that the reinforcement learning method gave better performance based on both wrapper evaluation and filter evaluation methods.

\section{Conclusion}
Based on our experimental results above, we are able to tune the hyper-parameter - epsilon to increase the agents exploration of states. We also evaluate the result of classification accuracy based on the selected dataset. However, we are not able to compare this with other methods directly since our method select the best subset and evaluates the accuracy based on this. This can only be compared with other embedded methods in future to verify that the reinforcement learning agent reached a state that has maximum classification accuracy based on the provided dataset. Comparing this method with other filter based methods in feature ranking, the RL agent achieved better feature ranking accuracy compared to other methods. 

\section{Future Work}
We hope to extend this model and improve its performance using more advanced Reinforcement Learning techniques such as function approximation so as to estimate the values of unseen states based on the values learned from the visited states. Another area we would consider for future research work is to apply policy iteration techniques such as Q-learning and SARSA combined with filter based methods so as to develop an optimal policy for the agent and by this control the actions that will be selected in every state.

\bibliographystyle{unsrt}  
\bibliography{references}



\end{document}